\documentclass[11pt]{article}

\usepackage[preprint]{acl}
\usepackage{booktabs} 
\usepackage{times}
\usepackage{float}
\usepackage{latexsym}
\usepackage{multirow}
\usepackage{subcaption}
\usepackage[utf8]{inputenc}
\usepackage[T2A,T1]{fontenc}
\DeclareFontFamily{T2A}{ptm}{}
\DeclareFontShape{T2A}{ptm}{m}{n}{<->ssub * cmr/m/n}{}
\usepackage[ukrainian,english]{babel}
\makeatletter
\renewcommand{\verbatim@font}{%
  \fontencoding{T2A}%
  \fontfamily{cmtt}%
  \selectfont
}
\usepackage{fvextra}

\DefineVerbatimEnvironment{PromptBlock}{Verbatim}{
  breaklines=true,
  breakanywhere=true,
  breaksymbolleft={},
  breaksymbolright={},
  fontsize=\footnotesize,
  formatcom=\fontencoding{T2A}\fontfamily{cmtt}\selectfont,
  frame=single,
  framesep=2mm
}
\makeatother
\usepackage{amsmath}

\usepackage{microtype}

\usepackage{inconsolata}

\usepackage{graphicx}

\title{UkrMeme: Multimodal Explainable Persuasion Detection in Ukrainian Memes}

\author{
  \textbf{Nazar Andrushko\textsuperscript{1}},
  \textbf{Vera Schmitt\textsuperscript{2,3,4}},
  \textbf{Veronika Solopova\textsuperscript{2,3}}
\\
\\
  \textsuperscript{1}Ukrainian Catholic University, Lviv, Ukraine \\
  \textsuperscript{2}Johannes Gutenberg University Mainz, Mainz, Germany\\
\textsuperscript{3}German Research Center for Artificial Intelligence (DFKI), Berlin, Germany\\
\textsuperscript{4}Centre for European Research in Trusted AI (CERTAIN)
\\
  \small{
  \texttt{nazar.checkemail@ucu.edu.ua} \quad
  \texttt{vera.schmitt@uni-mainz.de} \quad
  \texttt{veronika.solopova@uni-mainz.de}
  }
}

\hypersetup{pdftitle={UkrMeme: Multimodal Explainable Persuasion Detection in Ukrainian Memes}, pdfauthor={Nazar Andrushko, Vera Schmitt, Veronika Solopova}}

\begin{document}
\maketitle
\begin{abstract}
Memes play a central role in modern online communication, especially in contexts of political conflict where visual and textual cues combine to convey persuasive messages. We study persuasive technique detection in Ukrainian memes using vision–language models (VLMs). We introduce UkrMeme, a dataset of 200 organically produced Ukrainian wartime memes annotated for persuasion techniques, and PTM-UA, a Ukrainian translation of a subset of the Propaganda 950 benchmark that preserves original visual content while translating meme text. We evaluate VLMs in a zero-shot prompting setup across image, text, and multimodal inputs using hierarchical F1 metrics. Finally, we compare Layer-wise Relevance Propagation (LRP) and model-generated token attributions and analyze the semantic categories of evidence emphasized by each explanation method. 
\end{abstract}

\section{Introduction}

Memes have become one of the central communicative form in modern information ecosystem, combining visual and textual signals into compact multimodal artifacts. In contexts of political conflict and war, memes often serve as weapon of soft propaganda, especially easy to plant for foreign audiences, but it can also be instrument of collective sense-making, identity construction, and psychological coping \cite{LIAGUSHA2025101247,nato_memetic_warfare,10.1145/3711037}. Since the beginning of the full-scale invasion of Ukraine in 2022, Ukrainian online spaces have witnessed an unprecedented proliferation of war-related memes. Scholars and journalists have noted that Ukrainian meme culture functions as a form of grassroots resilience, black humor, and communal morale-building rather than merely as an instrument of coordinated information warfare \citep{reuters2023memes,bilaniuk2023memes,Bondarenko_2025,Bouchard28052025}.

While detection of persuasion techniques in multimodal content has recently received increased attention in Natural Language Processing (NLP), most existing benchmarks are English-centric and focus on identifying persuasive techniques in political messaging \citep{dimitrov-etal-2021-detecting}. Cross-lingual and culturally grounded analyses remain limited, especially for low-resource languages and contexts of active conflict. Moreover, the majority of prior work emphasizes model performance, with comparatively less attention to faithfulness and explainability of model decisions.

\begin{figure}[t]
    \centering
    \includegraphics[width=0.4\columnwidth]{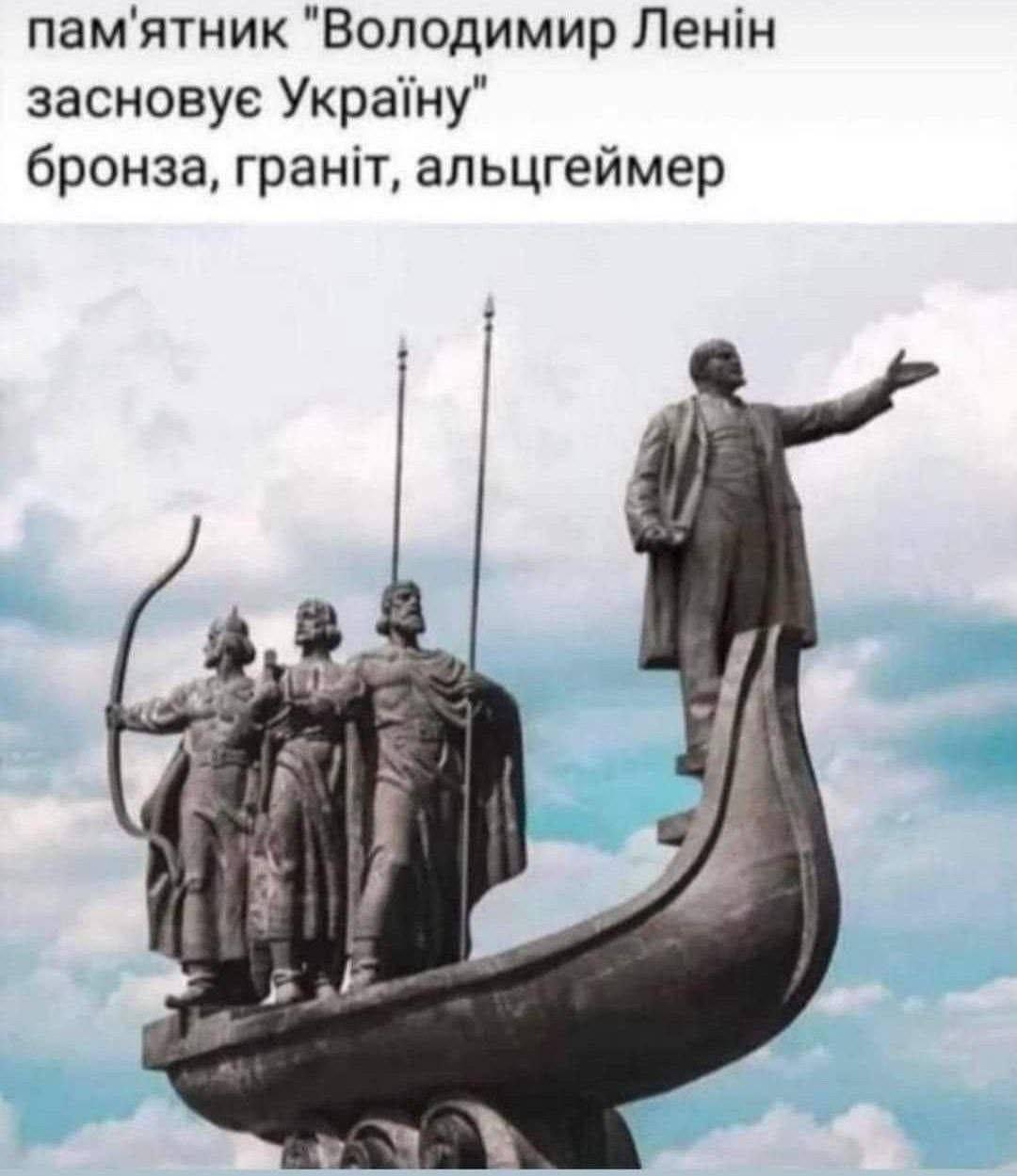}
    \caption{Example from the \textsc{UkrMeme} dataset. Gold labels: \textit{Smears}, \textit{Loaded Language}. en: "A monument to  “Vladimir Lenin founding Ukraine” bronze, granite, Alzheimer’s"}
    \label{fig:ukrmeme_sample}
\end{figure}

In this work, we investigate how large language models (LLMs) and vision-language models (VLMs) identify such techniques in memes in Ukrainian. Importantly, while we use persuasion labels from propaganda studies, this study does not assume that Ukrainian memes constitute state-coordinated propaganda, but a form of grassroots resilience. Hence, we treat persuasion techniques as an analytical lens for examining how persuasive rhetoric and emotional framing emerge organically in grassroots digital creativity during wartime. Our focus is on understanding both model behavior and the rhetorical structure of Ukrainian meme discourse. We construct two complementary datasets: (i) we translate into Ukrainian a subset of a widely used multimodal persuasive meme benchmark \cite{dimitrov-etal-2021-detecting}, preserving images while translating meme text into Ukrainian. This controlled setup allows us to isolate language effects while holding visual content constant; (ii) we compile and annotate a novel dataset of 200 organically produced Ukrainian wartime memes, curated from public online sources. Each meme is independently annotated by two native Ukrainian speakers using a multi-label taxonomy, followed by adjudication to ensure consistency. Figure~\ref{fig:ukrmeme_sample} shows a representative example from the dataset.

We evaluate multimodal VLM models (Qwen3-VL-4B-Instruct \cite{qwen3vl} and 8B version) and GPT-4.1-mini and Gemma 3 4B in a zero-shot prompting setup on both datasets. Beyond standard micro- and macro-averaged F1 scores, we conduct a faithfulness-oriented explainability analysis. Specifically, we compare token-level Layer-wise Relevance Propagation (LRP) \citep{bach2015lrp, montavon2019lrpreview} saliency with model based token attributions. We quantify overlap between these signals and analyze divergences across datasets and modalities.
Finally, we offer a pilot empirical perspective on how grassroots wartime meme culture in Ukraine employs persuasive techniques as part of civic resilience.

Both datasets and code is available in GitHub\footnote{\url{https://github.com/eingrid/persuasion-techniques-ua-memes}}.

\section{Related Work}

Automatic detection of persuasion techniques has been widely studied in textual political discourse. Early work focused on news articles and political speeches, introducing fine-grained taxonomies of persuasive techniques and corresponding supervised benchmarks \citep{da2019finegrained, dmitriev2019survey}. The labels and annotation quality for this task has been disputed as subjective and usually low-annotator agreement is expected \cite{sahitaj-etal-2025-hybrid}.
With the rise of multimodal communication, attention has shifted toward propaganda in internet memes, where meaning emerges from the interaction between text and image \citep{dimitrov-etal-2021-detecting, kmainasi-etal-2025-memeintel, memeperil}. These works demonstrate much lower performance than purely textual propaganda detection. At the same time, multimodal models were shown to outperform text-only systems on meme classification tasks, particularly when OCR-extracted text is incorporated into the visual pipeline. Notably, multimodal task inherited the labels from early propaganda works \cite{da-san-martino-etal-2019-fine}, with small image and cross-modality based label additions. 
However, existing benchmarks remain predominantly English-centric, while other languages and cross-lingual transfer in memes remain underexplored, especially in culturally specific contexts such as wartime meme production.
 
To our knowledge, no prior work systematically evaluates persuasion technique detection in Ukrainian memes under a prompt-only setting or compares text-only and vision-language models using faithfulness-oriented explainability metrics.
\section{Methods}
\subsection{Data}

We utilize two distinct datasets for our experiments: a translated subset of the Propaganda 950 dataset from ~\citet{dimitrov-etal-2021-detecting},  \textsc{PTM-UA} (Propaganda Techniques in Meme-Ukrainian) and a newly collected dataset of Ukrainian memes, \textsc{UkrMeme}.

\subsubsection{Translated Propaganda 950 (PTM-UA)}
To obtain a Ukrainian-language evaluation set, we translate the test split of the Propaganda 950 dataset~\cite{dimitrov-etal-2021-detecting} using a semi-automated pipeline. The process consists of four stages: (1) text detection using EasyOCR\footnote{\url{https://github.com/JaidedAI/EasyOCR}}, (2) removal of detected text via LaMa inpainting~\cite{suvorov2021resolution}\footnote{\url{https://github.com/enesmsahin/simple-lama-inpainting}}, (3) translation of the extracted English text into Ukrainian with GPT-5~\cite{OpenAI_GPT5_2025}, prompted to preserve rhetorical intent while producing concise, colloquial output, and (4) rendering the translated text back onto the image using automatic font sizing, text wrapping, and adaptive color selection for readability.

After the automated pipeline, we manually review all 200 translated test samples. We discard samples where the text placement was unsatisfactory (e.g., text overflowing bounding boxes or overlapping with important visual elements), where the inpainting process removed crucial visual information, or where the translation was inaccurate. The majority of rejected samples suffered from poor text placement, with a smaller number lost to inpainting artifacts and very few due to translation errors. After filtering, we retain \textbf{158 samples} for evaluation. The label distribution for the translated dataset is shown in Figure~\ref{fig:label_dist_translated}. Note that although the original Propaganda 950 dataset does not include an explicit No Propaganda label, a small number of test samples had none of the 22 propaganda techniques identified; we treat these as No Propaganda. 

\begin{figure}[t]
    \centering
    \includegraphics[width=0.7\columnwidth]{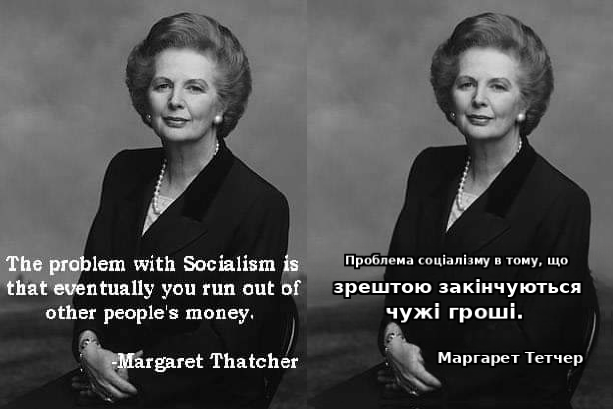}
    \caption{Example from PTM-UA: original English meme (left) and its Ukrainian translation (right). Text is inpainted and re-rendered while preserving the visual content. Gold labels: \textit{Appeal to Authority}, \textit{Smears}.}
    \label{fig:translated_sample}
\end{figure}

An example of the translation pipeline output is shown in Figure~\ref{fig:translated_sample}.

\subsubsection{\textsc{UkrMeme}}
 
To complement the translated data with natively Ukrainian content, we collect 200 memes from open Ukrainian-language sources, including Telegram channels and the SUCHO Meme Wall\footnote{\url{https://memes.sucho.org/}}. The images were collected in early 2026 and represent a variety of persuasive and non-persuasive content circulating in Ukrainian online spaces.
 
\begin{figure}[t]
    \centering
    \includegraphics[width=\columnwidth]{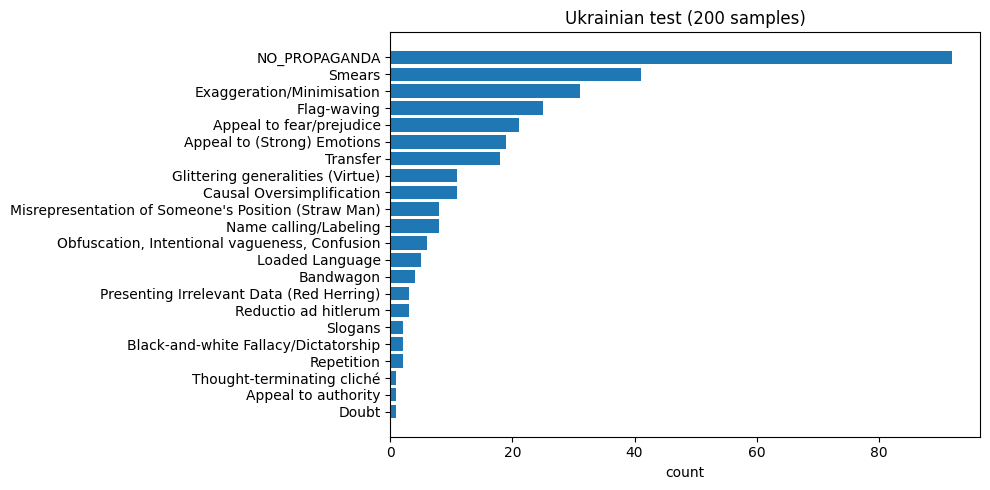}
    \caption{Label distribution for the \textsc{UkrMeme} dataset (200 samples).}
    \label{fig:label_dist_ukr}
\end{figure}

\begin{figure}[t]
    \centering
    \includegraphics[width=\columnwidth]{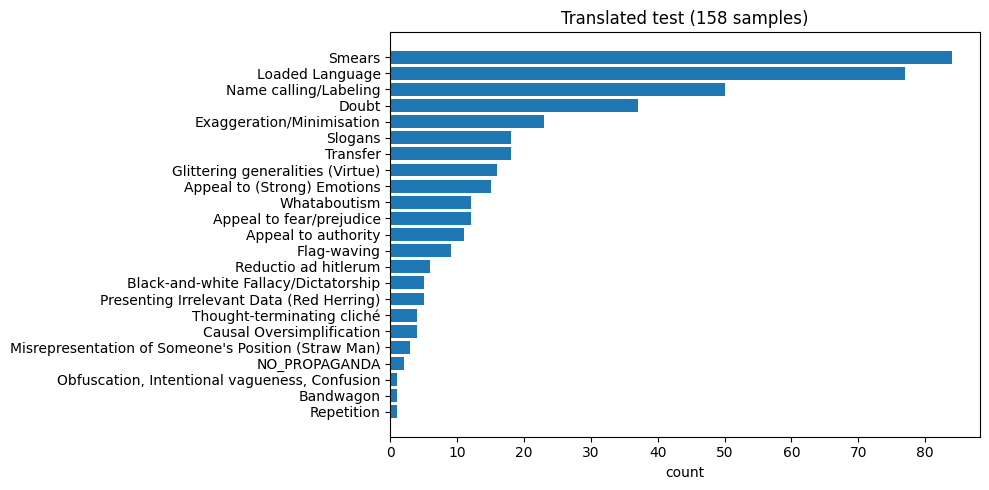}
    \caption{Label distribution for the translated PTM-UA (158 samples).}
    \label{fig:label_dist_translated}
\end{figure}

Two trained native-speaking annotators independently label each image using the same 22 technique labels as in the Propaganda 950 dataset~\cite{dimitrov-etal-2021-detecting}, plus a No Propaganda label\footnote{While we use "No Propaganda" label inheriting from previous datasets in the field, we underline that we do not assume that the rest of the dataset constitutes propaganda, but instead samples of  grassroots resilience memes.} for images that do not employ any persuasion technique. Disagreements are resolved by a designated curator. The label distribution (Figure~\ref{fig:label_dist_ukr}) shows a substantially larger share of No Propaganda samples and a different technique ranking compared to PTM-UA (Figure~\ref{fig:label_dist_translated}), reflecting the distinct nature of Ukrainian wartime meme content.
 
We report inter-annotator agreement following the multi-label methodology of \citet{marchal-etal-2022-establishing}. Table~\ref{tab:iaa_set} presents set-level measures; Table~\ref{tab:iaa} reports per-class Cohen's $\kappa$ for classes with at least 5 occurrences from either annotator. The MASI-weighted \cite{passonneau2006measuring} Krippendorff's $\alpha$ is \textbf{0.406} and the mean per-class $\kappa$ is \textbf{0.404}, indicating moderate agreement. Agreement is highest for \textit{No Propaganda} ($\kappa = 0.776$); among techniques, \textit{Smears} ($\kappa = 0.511$) and \textit{Appeal to Fear/Prejudice} ($\kappa = 0.506$) show the strongest agreement. Low $\kappa$ for classes such as \textit{Loaded Language} ($\kappa = 0.149$) reflects very low prevalence rather than poor annotation quality. Prior work attributes low agreement in this task to the inherent subjectivity of persuasion technique annotation and treats IAA as a signal for annotator management rather than a strict quality threshold \cite{sahitaj-etal-2025-hybrid}.

\begin{table}[t]
\footnotesize
\centering
\begin{tabular}{lccc}
\toprule
\textbf{Measure} & \textbf{Observed} & \textbf{Expected} & \textbf{Adjusted} \\
\midrule
Boot-match & 0.764 & 0.303 & 0.661 \\
Boot-precision & 0.628 & 0.184 & 0.544 \\
Boot-recall & 0.698 & 0.227 & 0.610 \\
Boot-F1 & 0.643 & 0.180 & 0.565 \\
Augmented-$\kappa$ & 0.535 & 0.238 & 0.390 \\
\midrule
\multicolumn{4}{l}{MASI Krippendorff's $\alpha$: \textbf{0.406}} \\
\bottomrule
\end{tabular}
\caption{Set-level inter-annotator agreement for \textsc{UkrMeme}.}
\label{tab:iaa_set}
\end{table}

\begin{table}[t]
\footnotesize
\centering
\begin{tabular}{lr}
\toprule
\textbf{Label} & \textbf{$\kappa$} \\
\midrule
No Propaganda & 0.776 \\
Smears & 0.511 \\
Appeal to Fear/Prejudice & 0.506 \\
Exaggeration/Minimisation & 0.415 \\
Causal Oversimplification & 0.409 \\
Glittering Generalities (Virtue) & 0.394 \\
Flag-waving & 0.358 \\
Name Calling/Labeling & 0.338 \\
Appeal to (Strong) Emotions & 0.305 \\
Transfer & 0.279 \\
Loaded Language & 0.149 \\
\midrule
\textbf{Mean} & \textbf{0.404} \\
\bottomrule
\end{tabular}
\caption{Per-class Cohen's $\kappa$ for \textsc{UkrMeme} labels with at least 5 occurrences from either annotator.}
\label{tab:iaa}
\end{table}

\subsection{Experimental set-up}

In this section, we describe our approach to persuasion detection and the explainability analysis we conduct using two different methods: Layer-wise Relevance Propagation (LRP) \cite{bach2015lrp} and VLM-based token attribution.

\subsubsection{Motivation}
LRP and VLM-based token attribution represent fundamentally different explainability approaches. LRP propagates relevance scores backward through the network, assigning importance based on each token's mathematical contribution to the output, a model-internal, gradient-based explanation. However, LRP explanations depend strongly on the chosen propagation rules and model architecture, meaning that analysis may reflect methodological choices rather than the model’s true decision process. Token attribution for Language models relies on the model's own reasoning to identify important tokens as part of its generation, which is a linguistic, model-external explanation. We hypothesize that comparing these complementary explanation methods can reveal systematic differences in the evidence emphasized by each approach.

\subsubsection{Models}
\label{sec:setup-models}

We evaluate four models across two tiers. \textbf{Qwen3-VL-4B-Instruct} and \textbf{Qwen3-VL-8B-Instruct} \cite{qwen3vl} are used for both persuasion detection and explainability analysis: as open-weight models, they provide full access to internal activations, enabling LRP alongside VLM-based token attribution on the same model. Comparing the two sizes additionally reveals scale effects on detection and explanation quality. We also report detection-only results for two additional models: \textbf{GPT-4.1-mini} and \textbf{Gemma 3 4B}. These are excluded from the explainability analysis for different reasons detailed in Appendix \ref{exclusion}.
\subsubsection{Detection Metrics}

We report hierarchical F1 following \cite{kiritchenko2006,dimitrov-etal-2024-semeval}, which awards partial credit when predicted and gold labels share ancestors in the taxonomy. The taxonomy organizes 22 techniques under categories such as Ethos, Pathos, Logos, and Ad Hominem \cite{dimitrov-etal-2024-semeval}. Each label set is extended to include all ancestor nodes except the root (Persuasion), and precision, recall, and F1 are computed over these extended sets. We report three variants: micro-averaged (the official SemEval metric), macro-averaged over taxonomy nodes, and mean per-sample F1. Full definitions are provided in Appendix~\ref{app:metrics}.

\subsubsection{Baselines and translation control}
We evaluate a weighted-random baseline to contextualize model performance. For each persuasion label, a binary prediction is sampled independently according to its empirical positive frequency in the dataset, and results are averaged over 30 runs. We report the same hierarchical micro- and macro-F1 metrics as for the evaluated models.

Additionally, to assess whether performance on PTM-UA is affected by translation into Ukrainian, we evaluate all four models on the original English versions of the same 158 Propaganda 950 samples retained in PTM-UA after filtering. We use identical input modalities and prompting settings, enabling a paired comparison in which meme identity and visual content are held constant while the textual condition differs between the original English and translated Ukrainian versions.

\subsubsection{Explanation Setup}
Since our datasets consist of visual content (images containing text), we evaluate across three input modes: (i) \textit{image}, where only the image is provided and the model must read text from it; (ii) \textit{image+text}, where both the image and its extracted text are provided in the prompt; and (iii) \textit{text}, where only the extracted text is provided without the image. This allows us to assess how input modality affects both detection performance and the resulting explanations. 
\paragraph{Prompting.} We use two zero-shot prompts: one for the image mode and another shared by the image+text and text modes. Both prompts provide the full set of persuasion technique classes and instruct the model to return a JSON object with three fields: \texttt{labels} (predicted persuasion techniques), \texttt{referred\_words} (individual words from the meme text most relevant to the classification), and \texttt{referred\_image\_concepts} (up to five visual elements supporting the classification, ranked by importance). Full prompts are provided in Appendix~\ref{app:prompts}. While the prompt also elicits image-level concept descriptions, we restrict our explainability analysis to textual token attributions due to the difficulty of systematically comparing free-text concept outputs with LRP spatial heatmaps.
\paragraph{LRP extraction.} We use \texttt{lxt.efficient} with \texttt{zennit} to monkey-patch the linear, gated-MLP, RMSNorm, LayerNorm, and attention layers of the Qwen3-VL text decoder with AttnLRP propagation rules \citep{pmlr-v235-achtibat24a} (library defaults, unmodified), replacing plain gradients in the backward pass. For each sample, we re-run a forward pass from the cached generated token sequence, with gradients enabled on the input embeddings, and set the propagation target as the sum of logits at the predicted label token(s). Relevance is the embedding times its gradient, summed over the embedding dimension, then normalized per sample by its max absolute value (non-finite values clipped to zero). For \textit{image+text}, this backward pass uses text tokens only; visual features are not re-injected. To obtain word-level scores, we align the OCR text with the decoded prompt via tokenizer offsets, assign each overlapping subword token to the word containing its span midpoint, and average absolute relevance per word. Words are filtered (stopwords, punctuation, length 1), and the number of LRP words kept per sample is capped to match the count of words in the model's own verbalized explanation for that sample (max $k=5$), for comparison.
\paragraph{Word-level comparison.} We compare VLM-attributed and LRP-selected words using Jaccard similarity to assess the agreement between the two explanation methods.
\paragraph{Category analysis.} To characterize the types of explanations each method produces, we categorize tokens into five categories: \textsc{Entity} (named people, places, organizations, brands), \textsc{Emotional} (emotionally loaded or fear/hate/outrage-inducing language), \textsc{Rhetoric} (calls to action, imperatives, slogans, persuasive framing), \textsc{Content} (neutral factual nouns, events, dates, statistics), and \textsc{Other} (numbers, punctuation artifacts, unclear items) using GPT-4o~\cite{openai2024gpt4o}. We then compare the category distributions across methods. The categorization prompt is provided in Appendix~\ref{app:category-prompt}, and its accuracy was validated on a manually annotated sample.

\section{Results}
\label{sec:results}

We report hierarchical F1 scores across two datasets (PTM-UA and UkrMeme), three input modalities (text, image, image+text) and four models. We also include a weighted-random baseline as a lower bound. For each label, this baseline draws a binary prediction independently with probability equal to that label's positive frequency in the training data, averaged over 30 runs. On PTM-UA it reaches hierarchical micro F1 of $0.376 \pm 0.024$ and macro F1 of $0.149 \pm 0.018$, with micro precision $0.503 \pm 0.030$ and recall $0.300 \pm 0.021$. On UkrMeme it reaches micro F1 of $0.256 \pm 0.025$ and macro F1 of $0.107 \pm 0.016$, with micro precision $0.257 \pm 0.024$ and recall $0.256 \pm 0.026$. Both are well below every model result in Table~\ref{tab:hf1-comparison}, showing that model predictions carry real signal beyond label-frequency priors. The gap between the two baselines is itself telling: PTM-UA's baseline is precision-heavy while UkrMeme's is roughly balanced, the same precision-recall asymmetry we observe for the models themselves below, and a sign of UkrMeme's flatter, No-Propaganda-heavy label distribution compared to PTM-UA. Qwen3-VL-4B and Qwen3-VL-8B are evaluated on both detection and explainability. GPT-4.1-mini and Gemma 3 4B serve as detection-only baselines (see \S\ref{sec:setup-models}). Table~\ref{tab:hf1-comparison} compares all four models on hF1 only, and Table~\ref{tab:multimodal-results} in Appendix B details other metrics for the Qwen models.

\begin{table*}[t]
\footnotesize
\centering
\setlength{\tabcolsep}{5pt}
\renewcommand{\arraystretch}{0.95}
\begin{tabular}{lll cccc}
\toprule
& & & \multicolumn{4}{c}{\textbf{hF1}} \\
\cmidrule(lr){4-7}
\textbf{Dataset} & \textbf{Averaging} & \textbf{Mode} & \textbf{Qwen3-VL-4B} & \textbf{Qwen3-VL-8B} & \textbf{GPT-4.1-mini}$^\dagger$ & \textbf{Gemma 3 4B}$^\dagger$ \\
\midrule

\multirow{9}{*}{PTM-UA}
& \multirow{3}{*}{Micro}
 & text       & 0.440 & 0.416 & 0.556 & 0.498 \\
&  & image      & 0.483 & \textbf{0.524} & 0.592 & 0.549 \\
&  & image+text & \textbf{0.537} & 0.523 & \textbf{0.613} & \textbf{0.569} \\
\cmidrule(lr){2-7}
& \multirow{3}{*}{Macro}
 & text       & 0.195 & 0.198 & 0.255 & 0.255 \\
&  & image      & 0.219 & 0.257 & 0.350 & 0.297 \\
&  & image+text & \textbf{0.245} & \textbf{0.291} & \textbf{0.356} & \textbf{0.309} \\
\cmidrule(lr){2-7}
& \multirow{3}{*}{Per-sample}
 & text       & 0.367 & 0.351 & 0.517 & 0.449 \\
&  & image      & 0.437 & \textbf{0.480} & 0.549 & 0.531 \\
&  & image+text & \textbf{0.489} & 0.475 & \textbf{0.577} & \textbf{0.541} \\
\midrule

\multirow{9}{*}{UkrMeme}
& \multirow{3}{*}{Micro}
 & text       & 0.305 & 0.326 & 0.343 & 0.288 \\
&  & image      & \textbf{0.323} & 0.337 & 0.379 & 0.324 \\
&  & image+text & 0.323 & \textbf{0.342} & \textbf{0.384} & \textbf{0.334} \\
\cmidrule(lr){2-7}
& \multirow{3}{*}{Macro}
 & text       & 0.125 & 0.156 & 0.160 & 0.131 \\
&  & image      & 0.142 & 0.183 & 0.192 & 0.166 \\
&  & image+text & \textbf{0.143} & \textbf{0.193} & \textbf{0.198} & \textbf{0.168} \\
\cmidrule(lr){2-7}
& \multirow{3}{*}{Per-sample}
 & text       & \textbf{0.277} & \textbf{0.388} & 0.235 & 0.191 \\
&  & image      & 0.256 & 0.274 & 0.258 & 0.246 \\
&  & image+text & 0.258 & 0.276 & \textbf{0.270} & \textbf{0.251} \\
\bottomrule
\end{tabular}
\caption{hF1 comparison across all four models on PTM-UA ($n{=}158$) and UkrMeme ($n{=}200$). Best hF1 per averaging method is \textbf{bolded} within each dataset and averaging type. $^\dagger$GPT-4.1-mini and Gemma 3 4B are detection-only baselines; explainability analysis is not conducted for these models (see \S\ref{sec:setup-models}).}
\label{tab:hf1-comparison}
\end{table*}

\subsection{Performance and Cross-Dataset Gap}

On PTM-UA, the best micro hF1 is 0.537 (Qwen3-VL-4B, image+text) with per-sample F1 of 0.489. On UkrMeme, performance drops substantially: micro F1 peaks at 0.342 (Qwen3-VL-8B, image+text) and per-sample F1 at 0.388. Macro F1 remains below 0.30 on PTM-UA and below 0.20 on UkrMeme, reflecting poor coverage of infrequent techniques. This persistent gap stems from three factors: (i) UkrMeme's large share of \textit{No Propaganda} samples, which the model frequently mislabels; (ii) the translated memes originate from English-language discourse well represented in pretraining, whereas UkrMeme requires Ukrainian wartime and cultural knowledge; and (iii) precision-recall profiles are inverted, PTM-UA is precision-oriented (P\,=\,0.569 vs.\ R\,=\,0.508 for 4B image+text), while UkrMeme is recall-oriented (P\,=\,0.272 vs.\ R\,=\,0.396), indicating systematic over-prediction on culturally unfamiliar content.

The two additional baselines reinforce these trends. GPT-4.1-mini achieves the highest micro hF1 across both datasets: 0.613 on PTM-UA (image+text) and 0.384 on UkrMeme (image+text), outperforming both Qwen models on detection. Gemma 3 4B lags behind GPT-4.1-mini but exceeds the Qwen models on PTM-UA recall-heavy metrics (micro hF1 0.569 image+text), driven by aggressive over-prediction (high recall, lower precision). On UkrMeme, Gemma 3 4B's performance is similar to Qwen3-VL-4B (micro hF1 0.334 vs.\ 0.323). The cross-dataset gap likely reflects several interacting factors, including UkrMeme’s larger share of No Propaganda samples, differences in label distribution, and the greater cultural and contextual specificity of native wartime memes. As examined in §\ref{sec:lang-ablation}, translation into Ukrainian additionally affects the Qwen models, although this effect is almost negligible in GPT-4.1-mini and Gemma 3 4B.

\subsection{Language Effect: Original vs.\ Translated English}
\label{sec:lang-ablation}

To isolate the effect of Ukrainian translation from dataset difficulty, we additionally run all four models on the \emph{original English} versions of the same 158 PTM-UA memes retained after filtering (\S3.1.1), using identical prompts and modalities, and compare against their translated-set hF1 from Table~\ref{tab:hf1-comparison}. Table~\ref{tab:lang-ablation} reports micro and macro hF1 for both conditions.

\begin{table}[t]
\footnotesize
\centering
\setlength{\tabcolsep}{3.5pt}
\begin{tabular}{llcc}
\toprule
\textbf{Model} & \textbf{Mode} & \textbf{Orig. EN} & \textbf{Translated} \\
& & Mi / Ma hF1 & Mi / Ma hF1 \\
\midrule
\multirow{3}{*}{GPT-4.1-mini}
 & text       & 0.548 / 0.270 & 0.556 / 0.255 \\
 & image      & 0.615 / 0.362 & 0.592 / 0.350 \\
 & image+text & 0.603 / 0.347 & 0.612 / 0.356 \\
\midrule
\multirow{3}{*}{Gemma 3 4B}
 & text       & 0.495 / 0.257 & 0.498 / 0.255 \\
 & image      & 0.542 / 0.309 & 0.549 / 0.297 \\
 & image+text & 0.566 / 0.318 & 0.569 / 0.314 \\
\midrule
\multirow{3}{*}{Qwen3-VL-4B}
 & text       & 0.482 / 0.214 & 0.440 / 0.195 \\
 & image      & 0.578 / 0.280 & 0.483 / 0.219 \\
 & image+text & 0.577 / 0.308 & 0.537 / 0.245 \\
\midrule
\multirow{3}{*}{Qwen3-VL-8B}
 & text       & 0.479 / 0.238 & 0.416 / 0.198 \\
 & image      & 0.553 / 0.326 & 0.524 / 0.257 \\
 & image+text & 0.548 / 0.338 & 0.523 / 0.291 \\
\bottomrule
\end{tabular}
\caption{Original English vs.\ Ukrainian-translated PTM-UA, hierarchical micro/macro F1, on the 158 retained samples. Translated-condition values for Qwen3-VL-4B/8B are taken from Table~\ref{tab:hf1-comparison}; Mi = micro hF1, Ma = macro hF1.}
\label{tab:lang-ablation}
\end{table}

For GPT-4.1-mini and Gemma 3 4B, performance is broadly similar between original and translated conditions (differences within $\pm0.03$ hF1 across all modes), indicating translation itself introduces little additional degradation beyond the source dataset's inherent difficulty. The Qwen models show a more consistent gap in the opposite direction: both 4B and 8B score higher on original English than on translated Ukrainian across every mode (e.g.\ 4B image+text: 0.577 vs.\ 0.537; 8B image+text: 0.548 vs.\ 0.523), suggesting that the Qwen models are comparatively more sensitive to the translated Ukrainian condition.

\subsection{Effect of Input Modality}

On PTM-UA, image+text yields the best F1 for the 4B model across all averaging methods, confirming that supplying extracted text reduces OCR errors. The 8B model's image-only performance is competitive with image+text (micro 0.524 vs.\ 0.523), suggesting its stronger OCR partially closes the gap. Text-only consistently lags behind, indicating visual context carries non-trivial signal.

On UkrMeme, modality effects diverge by metric. Image and image+text modes achieve higher micro F1 through improved recall, but text-only yields the best per-sample F1 for both 4B (0.277) and 8B (0.388). The image modes over-predict labels, with higher recall coming at the cost of substantially lower precision, penalizing individual samples. The model interprets visual cues in wartime memes as additional evidence for persuasion techniques, leading to systematic over-labeling.

\subsection{Effect of Model Scale}

Scaling from 4B to 8B yields uneven gains. On PTM-UA, the 8B model improves image-only micro F1 (0.524 vs.\ 0.483) and achieves the best macro F1 (0.291), indicating better rare-technique coverage. On UkrMeme, the largest gain appears in text-only per-sample F1 (0.388 vs.\ 0.277, a 40\% relative improvement). In multimodal modes, gains are modest on both datasets.

\subsection{No Propaganda Label Ablation}
\label{sec:no-prop-ablation}
 
The original Propaganda 950 taxonomy does not include a \textit{No Propaganda} label; we add it to the prompt as an explicit option. In PTM-UA only 2 samples genuinely lack annotations, yet both models predict \textit{No Propaganda} for 31 samples in text-only mode. Removing this option (Table~\ref{tab:no-propaganda-ablation} in Appendix~\ref{app:additional-results}) yields the largest gains in text-only mode; image and image+text improvements are marginal. The 4B model is more responsive: of the 31 text-only \textit{No Propaganda} predictions, only 8 become empty when the label is removed, with the rest replaced by technique labels. The 8B model is more conservative (22 of 31 become empty rather than committing to a label).

\subsection{Per-Class Confusion Patterns}

\begin{figure*}[t]
    \centering
    \includegraphics[width=1\linewidth]{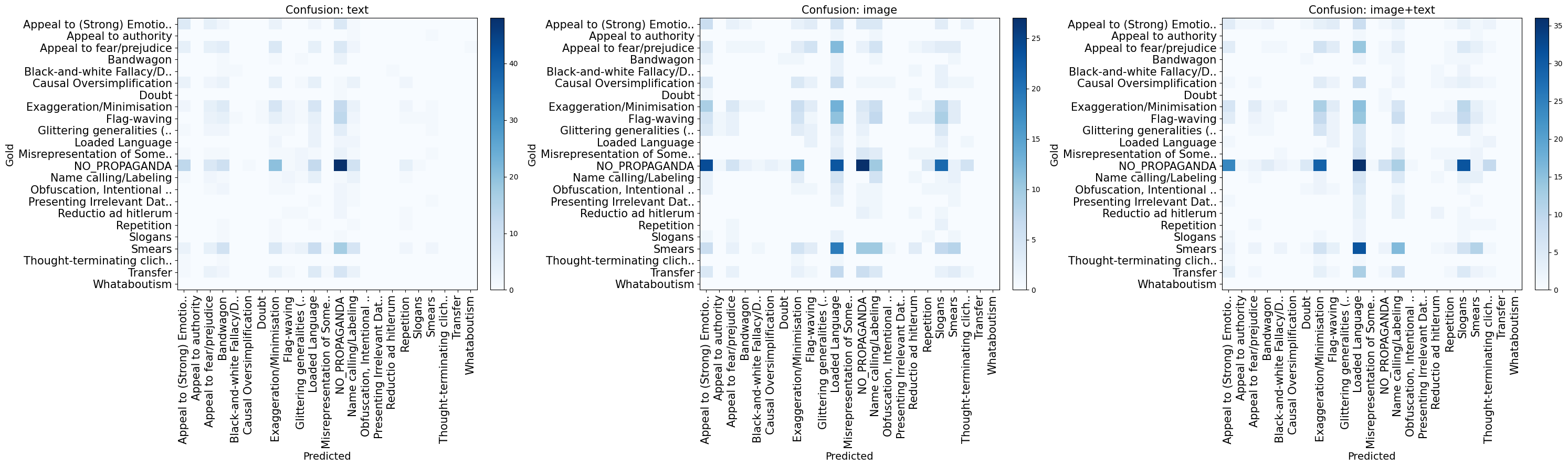}
    \caption{Confusion matrix for UkrMeme (Qwen3-VL-4B, image+text). The model massively over-predicts \textit{Loaded Language} and \textit{Slogans}, particularly for gold \textit{No Propaganda} samples.}
    \label{fig:cm_ukrmeme}
\end{figure*}

\begin{figure*}[t]
    \centering
    \includegraphics[width=1\linewidth]{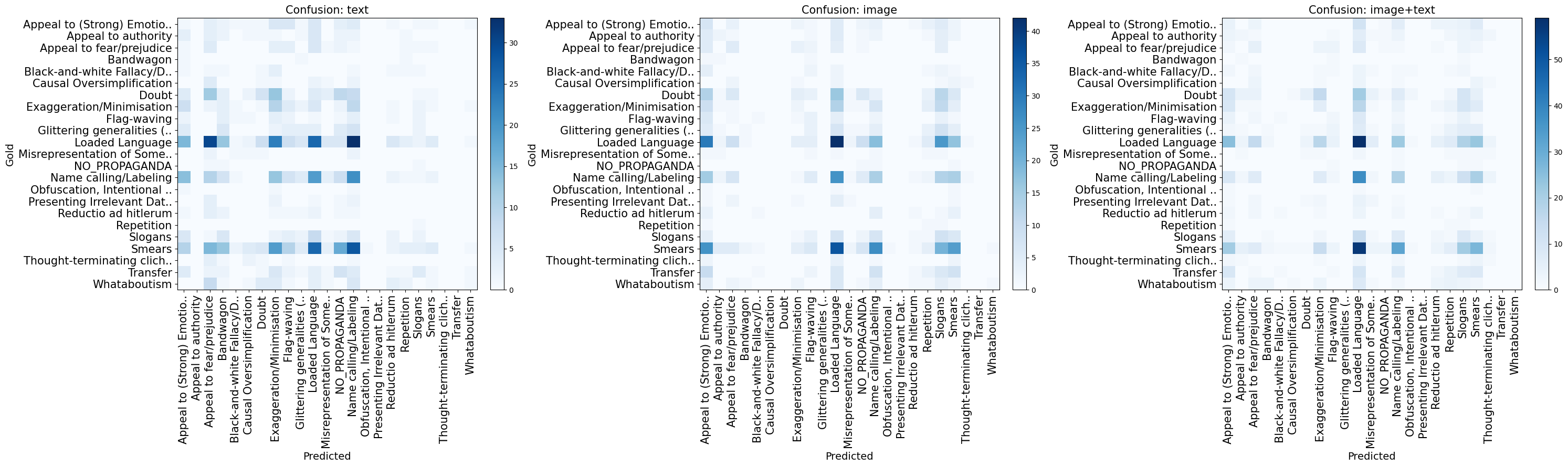}
    \caption{Confusion matrix for PTM-UA (Qwen3-VL-4B, image+text). \textit{Loaded Language} acts as the dominant attractor class.}
    \label{fig:cm_ptmua}
\end{figure*}

Figures~\ref{fig:cm_ukrmeme} and~\ref{fig:cm_ptmua} show confusion matrices for the 4B model across modalities. Cell $(i,j)$ is the raw count of samples where gold label $i$ and predicted label $j$ both occur: letting $G \in \{0,1\}^{N \times K}$ and $P \in \{0,1\}^{N \times K}$ be the binary gold and predicted label indicator matrices over $N$ samples and $K$ labels, the matrix is $G^\top P$; no row- or column-wise normalization is applied.\footnote{Since the task is multi-label, each cell counts gold-prediction label pairs; a sample with multiple gold and predicted labels contributes to several cells.} Figure~\ref{fig:cm_ptmua_noprop} in Appendix~\ref{app:additional-results} shows PTM-UA without the \textit{No Propaganda} label.

Across both datasets, \textit{Loaded Language} is the dominant attractor class: on PTM-UA it accumulates 283 column pairs against 214 row pairs, and on UkrMeme 179 column pairs against only 12 row pairs. The model defaults to this semantically broad label when uncertain. \textit{Smears} and \textit{Name Calling/Labeling} show strong mutual confusion on both datasets, consistent with their semantic proximity. Several mid-frequency classes (\textit{Black-and-white Fallacy}, \textit{Causal Oversimplification}, \textit{Glittering Generalities}, \textit{Transfer}) never appear on the diagonal in either dataset.

The key cross-dataset divergence concerns \textit{No Propaganda}: on PTM-UA this row contains only 4 label pairs, whereas on UkrMeme it contains 170 and drives the majority of false positives, 36 pairing with predicted \textit{Loaded Language}, 31 with \textit{Slogans}, and 29 with \textit{Exaggeration/Minimisation}. The model lacks a reliable mechanism for abstaining on emotionally charged but non-propagandistic wartime content. 

\subsection{LRP and token attribution}
\begin{figure}[t]
\centering

\begin{subfigure}[t]{0.48\linewidth}
    \centering
    \includegraphics[width=\linewidth]{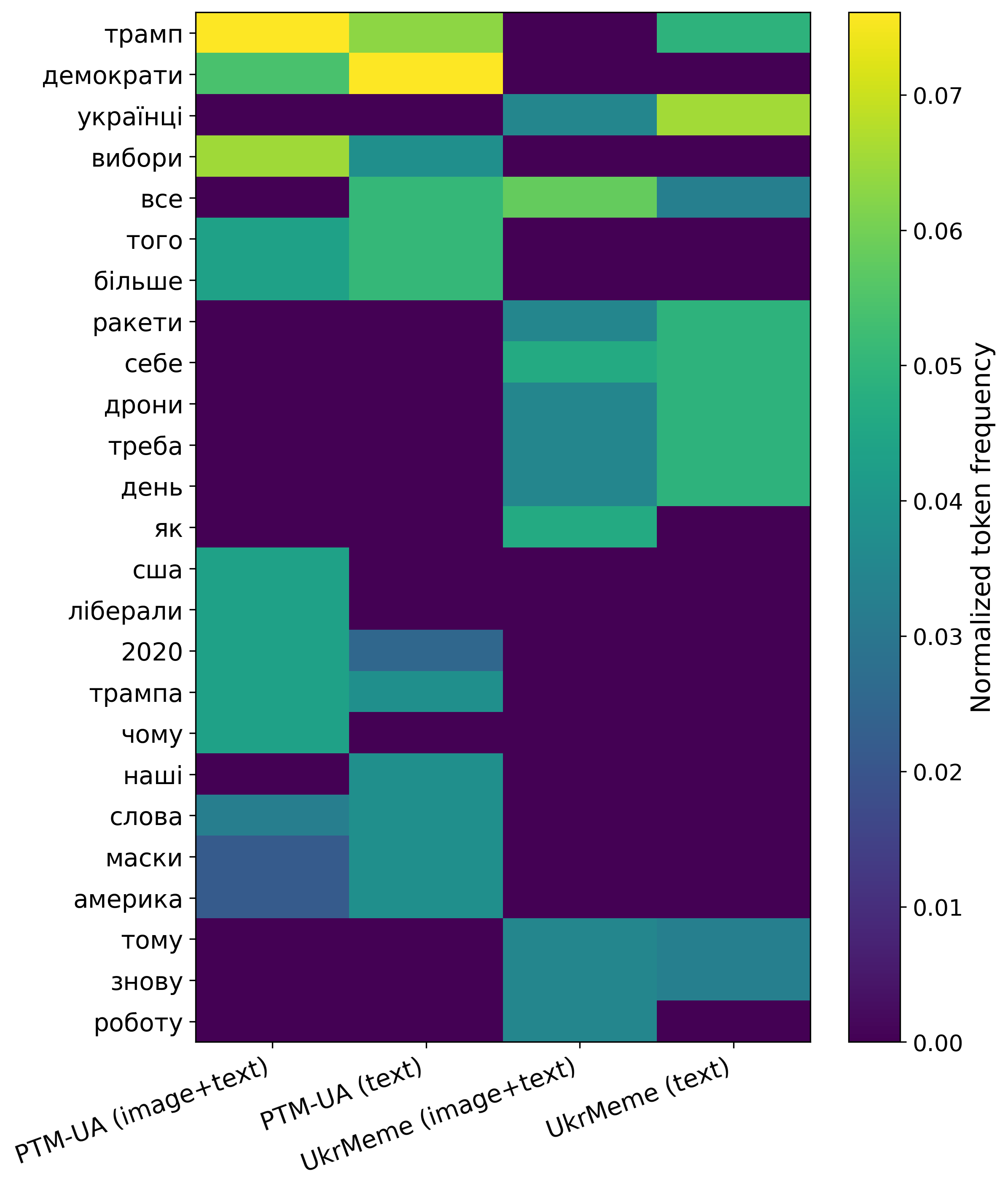}
    \caption{LRP attributions}
    \label{fig:lrp_heatmap}
\end{subfigure}
\hfill
\begin{subfigure}[t]{0.48\linewidth}
    \centering
    \includegraphics[width=\linewidth]{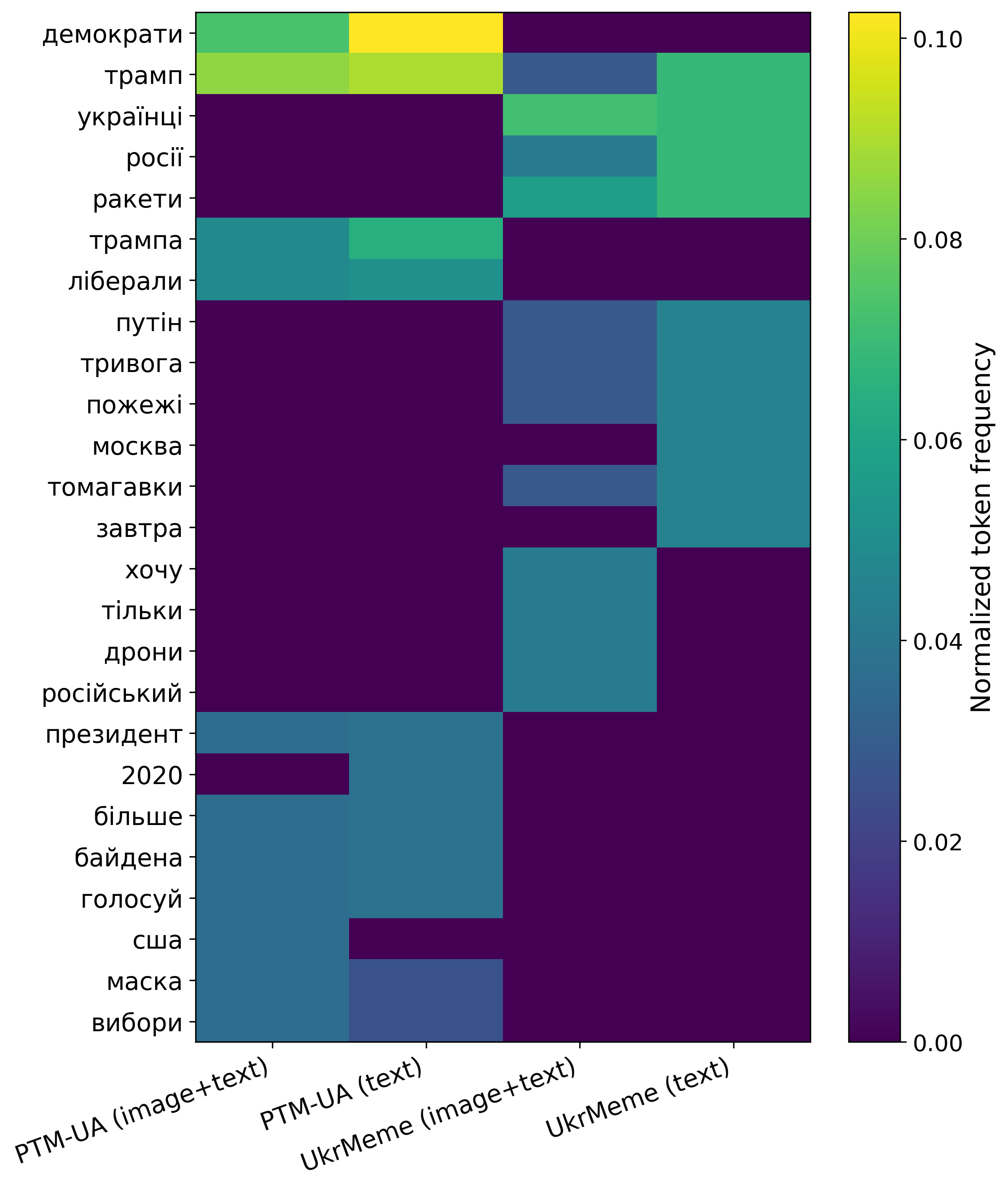}
    \caption{VLM attributions}
    \label{fig:llm_heatmap}
\end{subfigure}

\caption{Token attribution distributions across dataset settings. Columns correspond to PTM-UA (translated memes) and UkrMeme (native Ukrainian memes), with both multimodal (image+text) and text-only variants.}
\label{fig:token_heatmaps}

\end{figure}
\paragraph{Dataset related differences.} Across the four Qwen settings, the token-level explanations show a clear split between the translated and native Ukrainian meme conditions. In the translated dataset, both VLM-based token attribution and LRP are dominated by tokens tied to the original English political discourse (e.g. "trump", "democrats"). This pattern is stable for both text+image and text-only settings, suggesting that translation preserves the main lexical persuasion cues but also keeps the dataset anchored in the source-domain. In contrast, the native Ukrainian dataset is much more centered on the local wartime and sociopolitical context, with frequent tokens such as "ukrainians", "missiles", "drones", "Russia", "Putin", "alert", and "Tomahawk". 

Explanation agreement varies across dataset settings. Native multimodal memes exhibit the highest median overlap between VLM explanations and LRP attributions, suggesting more stable evidence selection. In contrast, translated memes show lower and more variable agreement, indicating that translation artifacts and multimodal context may destabilize token-level attribution alignment.

\paragraph{Modality differences.} The comparison between text+image and text-only also reveals an important modality effect. In the translated condition, the two modes are fairly similar, which suggests that much of the signal is recoverable from OCR text alone. In the native Ukrainian condition, however, the text+image model produces a broader and less purely lexical set of salient tokens, including more context-dependent tokens like "watch", "people", "works", "self", and "day", whereas the text-only model concentrates more on explicit textual triggers such as "missiles", "Trump", "Russia", and "Putin." This indicates that for native memes, the visual channel contributes additional contextual evidence that is not reducible to extracted text.
\begin{figure}[t]
\centering
\includegraphics[width=0.75\linewidth]{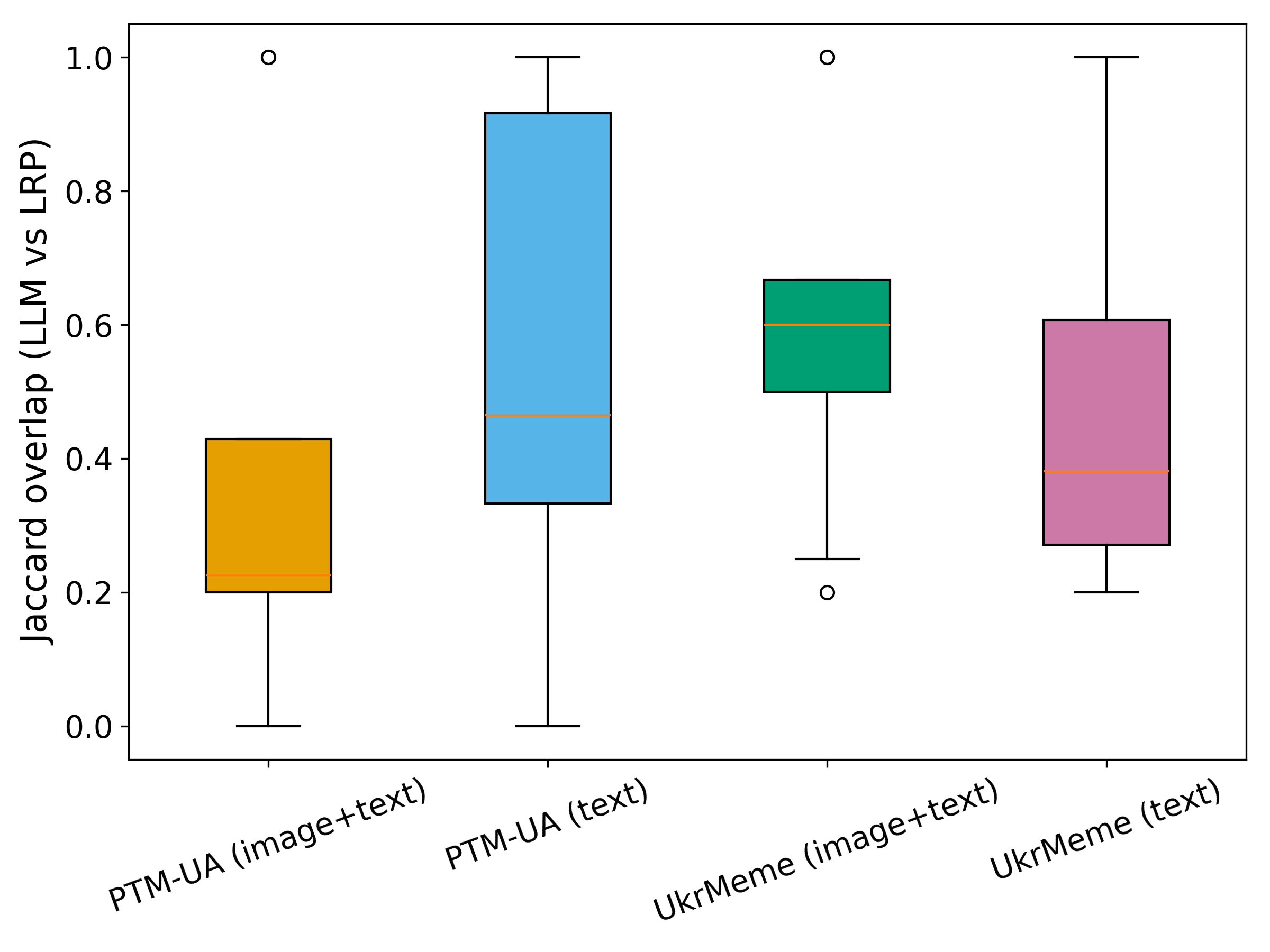}
\caption{Agreement between VLM-based and LRP token attributions explanations for low-hF1 samples across dataset settings. The distribution shows moderate overlap across conditions, with higher median agreement for the UkrMeme multimodal setting.}
\label{fig:agreement_boxplot}
\end{figure}
\paragraph{Low accuracy settings.} The low-hF1 (< 0.2) examples are especially informative. A recurring pattern is that VLM-based token attribution and LRP often overlap strongly even when the prediction is wrong, with several cases showing very high Jaccard agreement, even up to 1.0. That means the model is often consistently focusing on the same cues across explanation methods, but still mapping them to the wrong persuasion label. In other words, many failures are not due to random attention, but due to semantic confusion between nearby categories such as Smears, Loaded Language, Name-calling, Slogans, or Appeal to emotions. This is important because it suggests that the main weakness is not cue detection itself, but label discrimination.

At the same time, there are also low-hF1 cases with low overlap between VLM and LRP tokens, especially in some translated text-only examples. Those cases likely reflect OCR noise, translation artifacts, or unstable lexical grounding, where the model’s stated rationale and its internal relevance distribution diverge. This seems particularly visible in examples involving odd lexical fragments, borrowed English tokens, or noisy meme text.
\subsection{Category Analysis of Attributed tokens}
\label{sec:category-analysis}

We categorize all attributed tokens into five semantic categories (\textsc{Entity}, \textsc{Emotional}, \textsc{Rhetoric}, \textsc{Content}, \textsc{Other}; see Section~3) using GPT-4o \cite{openai2024gpt4o} with a structured categorization prompt (Appendix~\ref{app:category-prompt}), validated at 80.3\% accuracy on a manually annotated sample of 61 tokens. We partition tokens into \textit{shared} (selected by both methods), \textit{VLM-only}, and \textit{LRP-only} groups (Table~\ref{tab:category-source}).

\begin{table}[t]
\footnotesize
\centering
\setlength{\tabcolsep}{3.5pt}
\begin{tabular}{l rrrrrr}
\toprule
& \multicolumn{3}{c}{\textbf{PTM-UA}} & \multicolumn{3}{c}{\textbf{UkrMeme}} \\
\cmidrule(lr){2-4} \cmidrule(lr){5-7}
& \textbf{Sh.} & \textbf{VLM} & \textbf{LRP} & \textbf{Sh.} & \textbf{VLM} & \textbf{LRP} \\
\midrule
Entity    & 22.9 & 15.8 & 19.4 & 20.5 & 13.0 & 16.9 \\
Emotional & 33.1 & 33.7 & 12.7 & 25.9 & 32.4 & 12.7 \\
Rhetoric  & 17.1 & 22.7 & 19.4 & 15.5 & 17.6 & 15.9 \\
Content   & 23.4 & 26.4 & 34.3 & 33.2 & 30.9 & 33.1 \\
Other     &  3.6 &  1.1 & 14.2 &  4.9 &  6.2 & 21.3 \\
\midrule
$n$       & 363  & 273  & 268  & 425  & 324  & 314  \\
\bottomrule
\end{tabular}
\caption{Category distribution (\%) of attributed tokens by source. Total word counts shown as $n$.}
\label{tab:category-source}
\end{table}

The two methods select qualitatively different evidence. VLM-attributed tokens are disproportionately \textsc{Emotional} (33.7\% on PTM-UA, 32.4\% on UkrMeme) and \textsc{Rhetoric} (22.7\%, 17.6\%), suggesting that the model's verbalized reasoning gravitates toward affectively loaded and persuasive-sounding language when justifying its prediction.

LRP-only tokens show a different profile: \textsc{Content} dominates (${\sim}$34\% on both datasets), while \textsc{Emotional} drops to 12.7\%. The \textsc{Other} category is also elevated (14.2\% on PTM-UA, 21.3\% on UkrMeme), reflecting LRP's tendency to assign relevance to function words and punctuation that carry mathematical but not semantic importance.

Shared tokens have the highest \textsc{Entity} proportion (${\sim}$21-23\%) and lowest \textsc{Other} (${\sim}$4-5\%), indicating that when both methods converge, they agree on named entities, specific people, organizations, and places, as anchors for classification. These patterns are consistent across datasets despite strong differences in content and language, with the main divergence being a higher \textsc{Other} ratio in UkrMeme's LRP-only group, likely reflecting noisier OCR output in native Ukrainian memes.
\section{Discussion and Conclusion}
Our results highlight both the potential and limitations of current vision–language models for analyzing multimodal political discourse in low-resource and culturally specific contexts. While models achieve moderate performance on translated memes derived from English benchmarks, performance drops substantially on native Ukrainian memes, suggesting that cultural knowledge and contextual grounding remain key bottlenecks. Input modality also plays a complex role: multimodal inputs improve recall but often lead to systematic over-prediction of persuasion techniques, particularly in emotionally charged wartime content. Increasing model scale yields uneven gains, with improvements primarily in text-only settings rather than multimodal reasoning. Finally, our explainability analysis reveals a consistent divergence between model-based token attribution reasoning and LRP-based attributions: verbal explanations emphasize emotional and rhetorical cues, while internal relevance signals focus more on content words. Together, these findings underscore the challenges of reliable persuasion detection and interpretation in real-world multimodal settings.
\section*{Limitations}
Our study has several limitations. First, both datasets are relatively small: PTM-UA contains 158 translated memes and UkrMeme 200 native Ukrainian memes. While sufficient for exploratory evaluation, this scale limits statistical power and will not capture the full diversity of Ukrainian meme culture. Second, persuasion technique annotation is inherently subjective, and although inter-annotator agreement is moderate, disagreement reflects genuine ambiguity in rhetorical interpretation rather than purely annotation error, which is, however, conforming with reported agreement problem in prior work.

Third, the translated PTM-UA dataset preserves visual content from English memes while replacing text with Ukrainian translations. While this design isolates language effects, it may introduce translation artifacts and does not fully reproduce authentic Ukrainian meme discourse. Although originally motivated as a possible route toward data augmentation, our results suggest that simple translation-based augmentation is unlikely to provide substantial benefits. Moreover, the labels, coming from propaganda research, may not capture all of the discourse methods genuine civic society creation contains. Future work should analyse the memes in partnership with social scientists and linguists to identify genre-specific phenomena. 

Fourth, our experiments rely on zero-shot prompting and therefore do not evaluate supervised or fine-tuned systems. Fifth, explainability analysis is restricted to the two Qwen3-VL models: GPT-4.1-mini is excluded because LRP requires access to model-internal activations unavailable through a closed API, and Gemma 3 4B was excluded because running LRP over a 4B-parameter model for the full dataset was not computationally feasible within the scope of this study. Finally, our explainability analysis focuses on textual token attribution because visual concept explanations are difficult to compare systematically across methods. Future work should extend faithfulness analysis to spatial visual explanations, larger datasets, and open-weight models of comparable size to the closed-API baselines.
\section*{Ethical Considerations}
This work studies persuasion techniques in memes created during an ongoing armed conflict. Importantly, we do not assume that Ukrainian memes constitute coordinated propaganda; rather, we treat persuasion techniques as an analytical framework for examining rhetorical strategies in grassroots digital communication. Ukrainian wartime memes often serve functions of humor, resilience, and collective coping rather than strategic information warfare. While empirically our annotation procedure showed that persuasion techniques as a set of labels provided a sufficiently expressive annotation framework  encoded in UkrMeme dataset, with consideration of the sensitivity of the current war context, we recognize that some readers can find utilization of propaganda and persuasion techniques insulting. 

All memes in the UkrMeme dataset were collected from publicly accessible online sources. The dataset is used exclusively for research purposes and does not include private or personally identifiable information. Nevertheless, memes may contain politically sensitive or emotionally charged content. We therefore frame our analysis as descriptive rather than normative and caution against using automated persuasion detection systems as tools for censorship or political moderation. We acknowledge the broader risk that such research can be misused to suppress legitimate political expression. Our goal is to improve transparency and understanding of the phenomena, not to enable automated policing of political speech.
\section*{Acknowledgments}
This project has received funding from the European Union’s Horizon Europe research and innovation programme under Grant Agreement No. 101285890. Views and opinions expressed are however those of the author(s) only and do not necessarily reflect those of the European Union or the granting authority. Neither the European Union nor the granting authority can be held responsible for them.
\bibliography{custom}

@inproceedings{dimitrov-etal-2021-detecting,
    title = "Detecting Propaganda Techniques in Memes",
    author = "Dimitrov, Dimitar  and
      Bin Ali, Bishr  and
      Shaar, Shaden  and
      Alam, Firoj  and
      Silvestri, Fabrizio  and
      Firooz, Hamed  and
      Nakov, Preslav  and
      Da San Martino, Giovanni",
    editor = "Zong, Chengqing  and
      Xia, Fei  and
      Li, Wenjie  and
      Navigli, Roberto",
    booktitle = "Proceedings of the 59th Annual Meeting of the Association for Computational Linguistics and the 11th International Joint Conference on Natural Language Processing (Volume 1: Long Papers)",
    month = aug,
    year = "2021",
    address = "Online",
    publisher = "Association for Computational Linguistics",
    url = "https://aclanthology.org/2021.acl-long.516/",
    doi = "10.18653/v1/2021.acl-long.516",
    pages = "6603--6617"
}

@misc{OpenAI_GPT5_2025,
  author       = {{OpenAI}},
  title        = {{GPT-5 Large Language Model}},
  year         = {2025},
  url          = {https://chatgpt.com},
  note         = {Model version/snapshot from 07 Aug 2025}
}

@inproceedings{da2019finegrained,
  title={Fine-Grained Analysis of Propaganda in News Articles},
  author={Da San Martino, Giovanni and Yu, Seunghak and Barr{\'o}n-Cede{\~n}o, Alberto and Petrov, Dimitar and Nakov, Preslav},
  booktitle={Proceedings of EMNLP},
  year={2019}
}

@inproceedings{dmitriev2019survey,
  title={A Survey on Propaganda Detection},
  author={Dmitriev, Denis and others},
  booktitle={Proceedings of RANLP},
  year={2019}
}

@article{memeperil,
author = {Singh, Yashashvi and Kesharwani, Adwait and Chaturvedi, Animesh and Sharma, Rajesh},
year = {2025},
month = {11},
pages = {},
title = {MemePeril: a diverse multi-modal harmful memes dataset for the evolution analytics and LLM prompts},
volume = {3},
journal = {Discover Data},
doi = {10.1007/s44248-025-00090-y}
}

@inproceedings{kmainasi-etal-2025-memeintel,
    title = "{M}eme{I}ntel: Explainable Detection of Propagandistic and Hateful Memes",
    author = "Kmainasi, Mohamed Bayan  and
      Hasnat, Abul  and
      Hasan, Md Arid  and
      Shahroor, Ali Ezzat  and
      Alam, Firoj",
    editor = "Christodoulopoulos, Christos  and
      Chakraborty, Tanmoy  and
      Rose, Carolyn  and
      Peng, Violet",
    booktitle = "Proceedings of the 2025 Conference on Empirical Methods in Natural Language Processing",
    month = nov,
    year = "2025",
    address = "Suzhou, China",
    publisher = "Association for Computational Linguistics",
    url = "https://aclanthology.org/2025.emnlp-main.1539/",
    doi = "10.18653/v1/2025.emnlp-main.1539",
    pages = "30263--30279",
    ISBN = "979-8-89176-332-6"
}

@inproceedings{da-san-martino-etal-2019-fine,
    title = "Fine-Grained Analysis of Propaganda in News Articles",
    author = "Da San Martino, Giovanni  and
      Yu, Seunghak  and
      Barr{\'o}n-Cede{\~n}o, Alberto  and
      Petrov, Rostislav  and
      Nakov, Preslav",
    editor = "Inui, Kentaro  and
      Jiang, Jing  and
      Ng, Vincent  and
      Wan, Xiaojun",
    booktitle = "Proceedings of the 2019 Conference on Empirical Methods in Natural Language Processing and the 9th International Joint Conference on Natural Language Processing (EMNLP-IJCNLP)",
    month = nov,
    year = "2019",
    address = "Hong Kong, China",
    publisher = "Association for Computational Linguistics",
    url = "https://aclanthology.org/D19-1565/",
    doi = "10.18653/v1/D19-1565",
    pages = "5636--5646"
}

@inproceedings{sahitaj-etal-2025-hybrid,
    title = "Hybrid Annotation for Propaganda Detection: Integrating {LLM} Pre-Annotations with Human Intelligence",
    author = {Sahitaj, Ariana  and
      Sahitaj, Premtim  and
      Solopova, Veronika  and
      Li, Jiaao  and
      M{\"o}ller, Sebastian  and
      Schmitt, Vera},
    editor = "Atwell, Katherine  and
      Biester, Laura  and
      Borah, Angana  and
      Dementieva, Daryna  and
      Ignat, Oana  and
      Kotonya, Neema  and
      Liu, Ziyi  and
      Wan, Ruyuan  and
      Wilson, Steven  and
      Zhao, Jieyu",
    booktitle = "Proceedings of the Fourth Workshop on NLP for Positive Impact (NLP4PI)",
    month = jul,
    year = "2025",
    address = "Vienna, Austria",
    publisher = "Association for Computational Linguistics",
    url = "https://aclanthology.org/2025.nlp4pi-1.18/",
    doi = "10.18653/v1/2025.nlp4pi-1.18",
    pages = "215--228",
    ISBN = "978-1-959429-19-7"
}

@article{bach2015lrp,
  title={On Pixel-Wise Explanations for Non-Linear Classifier Decisions by Layer-Wise Relevance Propagation},
  author={Bach, Sebastian and Binder, Alexander and Montavon, Gr{\'e}goire and Klauschen, Frederick and M{\"u}ller, Klaus-Robert and Samek, Wojciech},
  journal={PLOS ONE},
  volume={10},
  number={7},
  pages={e0130140},
  year={2015},
  doi={10.1371/journal.pone.0130140}
}

@article{montavon2019lrpreview,
  title={Layer-Wise Relevance Propagation: An Overview},
  author={Montavon, Gr{\'e}goire and Binder, Alexander and Lapuschkin, Sebastian and Samek, Wojciech and M{\"u}ller, Klaus-Robert},
  journal={Explainable AI: Interpreting, Explaining and Visualizing Deep Learning},
  publisher={Springer},
  pages={193--209},
  year={2019},
  doi={10.1007/978-3-030-28954-6_10}
}

@misc{reuters2023memes,
  title={Memes and morale: decoding Ukraine’s comedy arsenal against disinformation},
  author={{Reuters Institute for the Study of Journalism}},
  year={2023},
  url={https://reutersinstitute.politics.ox.ac.uk/memes-morale-decoding-ukraines-comedy-arsenal-against-disinformation}
}

@incollection{bilaniuk2023memes,
  title={Memes as Antibodies: Humor and Resilience in Ukraine},
  author={Bilaniuk, Laada},
  booktitle={Language in the Time of War},
  publisher={Routledge},
  year={2023}
}

@article{Bondarenko_2025, title={Black humour in Ukrainian war memes: cognitive issues}, volume={13}, url={https://europeanjournalofhumour.org/ejhr/article/view/932}, DOI={10.7592/EJHR.2025.13.2.932}, abstractNote={
This article investigates the cognitive foundations of Ukrainian war-related black humour memes through their conceptual modeling. Black humour is reinterpreted as a meme genre through the lens of fundamental humour theories. The study’s objectives include, first, exploring the cultural conceptual context of Ukrainian black humour memes, and, second explaining the cognitive nature of memetic black humour by decomposing it into successive mental operations involving concept DEATH. To achieve these objectives, a case study of 100 randomly selected black humour memes from Ukrainian Telegram chats draws on the theory of national construal of the world, conceptual integration, and Extended Metaphor Theory. Within these theoretical frameworks, I model a complex construal. First, the study demonstrates that black humour memes are seamlessly integrated into the national construal of the world, a conceptual structure rooted in national values. Unlike ordinary humour representations, black humour memes paradoxically contribute to the national construal of the world by incorporating concepts or values traditionally avoided in nation building. Second, the article deconstructs a meme into several input spaces, a generic space where metaphorical mapping occurs, and a blend that conveys the meme’s message. Metaphorical mapping is selective as only certain elements of input spaces are transformed. The remaining salient elements shift into the blend untransformed. }, number={2}, journal={The European Journal of Humour Research}, author={Bondarenko, Ievgeniia}, year={2025}, month={Jul.}, pages={41–62} }

@article{Bouchard28052025,
author = {Michel Bouchard and Daria Antsybor},
title = {Memeing war: the use of humor for hope, resistance, and forging the nation},
journal = {National Identities},
volume = {0},
number = {0},
pages = {1--34},
year = {2025},
publisher = {Routledge},
doi = {10.1080/14608944.2025.2505477},


URL = { 
    
        https://doi.org/10.1080/14608944.2025.2505477
    
    

},
eprint = { 
    
        https://doi.org/10.1080/14608944.2025.2505477
    
    

}

}

@article{10.1145/3711037,
author = {Mejova, Yelena and Capozzi, Arthur and Monti, Corrado and De Francisci Morales, Gianmarco},
title = {Narratives of War: Ukrainian Memetic Warfare on Twitter},
year = {2025},
issue_date = {May 2025},
publisher = {Association for Computing Machinery},
address = {New York, NY, USA},
volume = {9},
number = {2},
url = {https://doi.org/10.1145/3711037},
doi = {10.1145/3711037},
journal = {Proc. ACM Hum.-Comput. Interact.},
month = may,
articleno = {CSCW139},
numpages = {28}
}

@misc{nato_memetic_warfare,
  title={It's time to embrace memetic warfare},
  author={NATO Strategic Communications Centre of Excellence},
  year={2015},
  note={defining memetic warfare as narrative and idea competition akin to propaganda},
  url={https://arxiv.org/html/2309.08363v3}
}

@article{LIAGUSHA2025101247,
title = {Memes, freedom, and resilience to information disorders: Information warfare between democracies and autocracies},
journal = {Social Sciences \& Humanities Open},
volume = {11},
pages = {101247},
year = {2025},
issn = {2590-2911},
doi = {https://doi.org/10.1016/j.ssaho.2024.101247},
url = {https://www.sciencedirect.com/science/article/pii/S2590291124004443},
author = {Anton Liagusha and Dmytro Iarovyi}
}

@inproceedings{suvorov2021resolution,
  title={Resolution-robust Large Mask Inpainting with {F}ourier Convolutions},
  author={Suvorov, Roman and Logacheva, Elizaveta and Mashikhin, Anton and Remizova, Anastasia and Ashukha, Arsenii and Silvestrov, Aleksei and Kong, Naejin and Goka, Harshith and Park, Kiwoong and Lempitsky, Victor},
  booktitle={Proceedings of the IEEE/CVF Winter Conference on Applications of Computer Vision (WACV)},
  year={2022}
}

@inproceedings{marchal-etal-2022-establishing,
    title = "Establishing Annotation Quality in Multi-label Annotations",
    author = "Marchal, Marian and Scholman, Merel and Yung, Frances and Demberg, Vera",
    booktitle = "Proceedings of the 29th International Conference on Computational Linguistics",
    month = oct,
    year = "2022",
    address = "Gyeongju, Republic of Korea",
    publisher = "International Committee on Computational Linguistics",
    url = "https://aclanthology.org/2022.coling-1.322/",
    pages = "3659--3668"
}

@inproceedings{passonneau2006measuring,
    title = "Measuring Agreement on Set-valued Items ({MASI}) for Semantic and Pragmatic Annotation",
    author = "Passonneau, Rebecca",
    booktitle = "Proceedings of the Fifth International Conference on Language Resources and Evaluation ({LREC})",
    year = "2006",
    url = "http://www.lrec-conf.org/proceedings/lrec2006/pdf/636_pdf.pdf"
}

@inproceedings{kiritchenko2006,
    title = "Learning and Evaluation in the Presence of Class Hierarchies: Application to Text Categorization",
    author = "Kiritchenko, Svetlana and Matwin, Stan and Nock, Richard and Famili, A. Fazel",
    booktitle = "Advances in Artificial Intelligence",
    series = "Lecture Notes in Computer Science",
    volume = "4013",
    pages = "395--406",
    year = "2006",
    publisher = "Springer"
}

@article{qwen3vl,
    title={Qwen3-VL Technical Report},
    author={{Qwen Team}},
    journal={arXiv preprint arXiv:2511.21631},
    year={2025}
}

@article{openai2024gpt4o,
    title={{GPT-4o} System Card},
    author={{OpenAI}},
    journal={arXiv preprint arXiv:2410.21276},
    year={2024}
}

@inproceedings{dimitrov-etal-2024-semeval,
    title = "{S}em{E}val-2024 Task 4: Multilingual Detection of Persuasion Techniques in Memes",
    author = "Dimitrov, Dimitar and Alam, Firoj and Hasanain, Maram and Hasnat, Abul and Silvestri, Fabrizio and Nakov, Preslav and Da San Martino, Giovanni",
    booktitle = "Proceedings of the 18th International Workshop on Semantic Evaluation (SemEval-2024)",
    month = jun,
    year = "2024",
    address = "Mexico City, Mexico",
    publisher = "Association for Computational Linguistics",
    pages = "2009--2026",
}

@InProceedings{pmlr-v235-achtibat24a,
  title = {{A}ttn{LRP}: Attention-Aware Layer-Wise Relevance Propagation for Transformers},
  author = {Achtibat, Reduan and Hatefi, Sayed Mohammad Vakilzadeh and Dreyer, Maximilian and Jain, Aakriti and Wiegand, Thomas and Lapuschkin, Sebastian and Samek, Wojciech},
  booktitle = {Proceedings of the 41st International Conference on Machine Learning},
  pages = {135--168},
  year = {2024},
  editor = {Salakhutdinov, Ruslan and Kolter, Zico and Heller, Katherine and Weller, Adrian and Oliver, Nuria and Scarlett, Jonathan and Berkenkamp, Felix},
  volume = {235},
  series = {Proceedings of Machine Learning Research},
  month = {21--27 Jul},
  publisher = {PMLR}
}
\onecolumn
\appendix
\section{Additional Results}
\label{app:additional-results}

\begin{table}[H]
\small
\centering
\setlength{\tabcolsep}{4pt}
\renewcommand{\arraystretch}{0.95}
\begin{tabular}{lll lll lll}
\toprule
& & & \multicolumn{3}{c}{\textbf{Qwen3-VL-4B}} & \multicolumn{3}{c}{\textbf{Qwen3-VL-8B}} \\
\cmidrule(lr){4-6} \cmidrule(lr){7-9}
\textbf{Dataset} & \textbf{Averaging} & \textbf{Mode} & hP & hR & hF1 & hP & hR & hF1 \\
\midrule

\multirow{9}{*}{PTM-UA}
& \multirow{3}{*}{Micro}
 & text       & 0.538 & 0.372 & 0.440 & 0.492 & 0.360 & 0.416 \\
&  & image      & 0.570 & 0.419 & 0.483 & 0.536 & 0.512 & \textbf{0.524} \\
&  & image+text & 0.569 & 0.508 & \textbf{0.537} & 0.519 & 0.526 & 0.523 \\
\cmidrule(lr){2-9}
& \multirow{3}{*}{Macro}
 & text       & 0.308 & 0.175 & 0.195 & 0.328 & 0.210 & 0.198 \\
&  & image      & 0.302 & 0.249 & 0.219 & 0.334 & 0.295 & 0.257 \\
&  & image+text & 0.280 & 0.277 & \textbf{0.245} & 0.354 & 0.320 & \textbf{0.291} \\
\cmidrule(lr){2-9}
& \multirow{3}{*}{Per-sample}
 & text       & 0.436 & 0.339 & 0.367 & 0.408 & 0.332 & 0.351 \\
&  & image      & 0.512 & 0.411 & 0.437 & 0.514 & 0.479 & \textbf{0.480} \\
&  & image+text & 0.538 & 0.480 & \textbf{0.489} & 0.492 & 0.492 & 0.475 \\
\midrule

\multirow{9}{*}{UkrMeme}
& \multirow{3}{*}{Micro}
 & text       & 0.272 & 0.346 & 0.305 & 0.343 & 0.310 & 0.326 \\
&  & image      & 0.271 & 0.399 & \textbf{0.323} & 0.284 & 0.413 & 0.337 \\
&  & image+text & 0.272 & 0.396 & 0.323 & 0.279 & 0.442 & \textbf{0.342} \\
\cmidrule(lr){2-9}
& \multirow{3}{*}{Macro}
 & text       & 0.123 & 0.193 & 0.125 & 0.187 & 0.190 & 0.156 \\
&  & image      & 0.150 & 0.209 & 0.142 & 0.195 & 0.289 & 0.183 \\
&  & image+text & 0.183 & 0.196 & \textbf{0.143} & 0.204 & 0.276 & \textbf{0.193} \\
\cmidrule(lr){2-9}
& \multirow{3}{*}{Per-sample}
 & text       & 0.294 & 0.277 & \textbf{0.277} & 0.408 & 0.385 & \textbf{0.388} \\
&  & image      & 0.264 & 0.266 & 0.256 & 0.276 & 0.296 & 0.274 \\
&  & image+text & 0.267 & 0.267 & 0.258 & 0.269 & 0.306 & 0.276 \\
\bottomrule
\end{tabular}
\caption{Hierarchical F1 results on PTM-UA ($n{=}158$) and UkrMeme ($n{=}200$) for Qwen3-VL models. Best hF1 per averaging method is \textbf{bolded} within each model and dataset.}
\label{tab:multimodal-results}
\end{table}
\begin{table}[H]
\small
\centering
\setlength{\tabcolsep}{4pt}
\begin{tabular}{ll lll lll}
\toprule
& & \multicolumn{3}{c}{\textbf{Qwen3-VL-4B}} & \multicolumn{3}{c}{\textbf{Qwen3-VL-8B}} \\
\cmidrule(lr){3-5} \cmidrule(lr){6-8}
\textbf{Averaging} & \textbf{Mode} & w/ NP & w/o NP & $\Delta$ & w/ NP & w/o NP & $\Delta$ \\
\midrule
\multirow{3}{*}{Micro}
 & text       & 0.440 & \textbf{0.475} & \textcolor{gray}{+.035} & 0.416 & \textbf{0.432} & \textcolor{gray}{+.016} \\
 & image      & 0.483 & \textbf{0.527} & \textcolor{gray}{+.044} & \textbf{0.524} & 0.515 & \textcolor{gray}{$-$.009} \\
 & image+text & 0.537 & \textbf{0.550} & \textcolor{gray}{+.013} & 0.523 & \textbf{0.529} & \textcolor{gray}{+.006} \\
\midrule
\multirow{3}{*}{Macro}
 & text       & 0.195 & \textbf{0.214} & \textcolor{gray}{+.019} & 0.198 & \textbf{0.211} & \textcolor{gray}{+.013} \\
 & image      & 0.219 & \textbf{0.229} & \textcolor{gray}{+.010} & 0.257 & \textbf{0.276} & \textcolor{gray}{+.019} \\
 & image+text & 0.245 & \textbf{0.249} & \textcolor{gray}{+.004} & 0.291 & \textbf{0.301} & \textcolor{gray}{+.010} \\
\midrule
\multirow{3}{*}{Per-sample}
 & text       & 0.367 & \textbf{0.427} & \textcolor{gray}{+.060} & 0.351 & \textbf{0.368} & \textcolor{gray}{+.017} \\
 & image      & 0.437 & \textbf{0.482} & \textcolor{gray}{+.045} & \textbf{0.480} & 0.470 & \textcolor{gray}{$-$.010} \\
 & image+text & 0.489 & \textbf{0.511} & \textcolor{gray}{+.022} & 0.475 & \textbf{0.485} & \textcolor{gray}{+.010} \\
\bottomrule
\end{tabular}
\caption{PTM-UA results with and without the \textit{No Propaganda} label ($n{=}158$). The largest improvements appear in text-only mode, where \textit{No Propaganda} is most frequently predicted.}
\label{tab:no-propaganda-ablation}
\end{table}
\begin{figure}[H]
    \centering
    \includegraphics[width=1\linewidth]{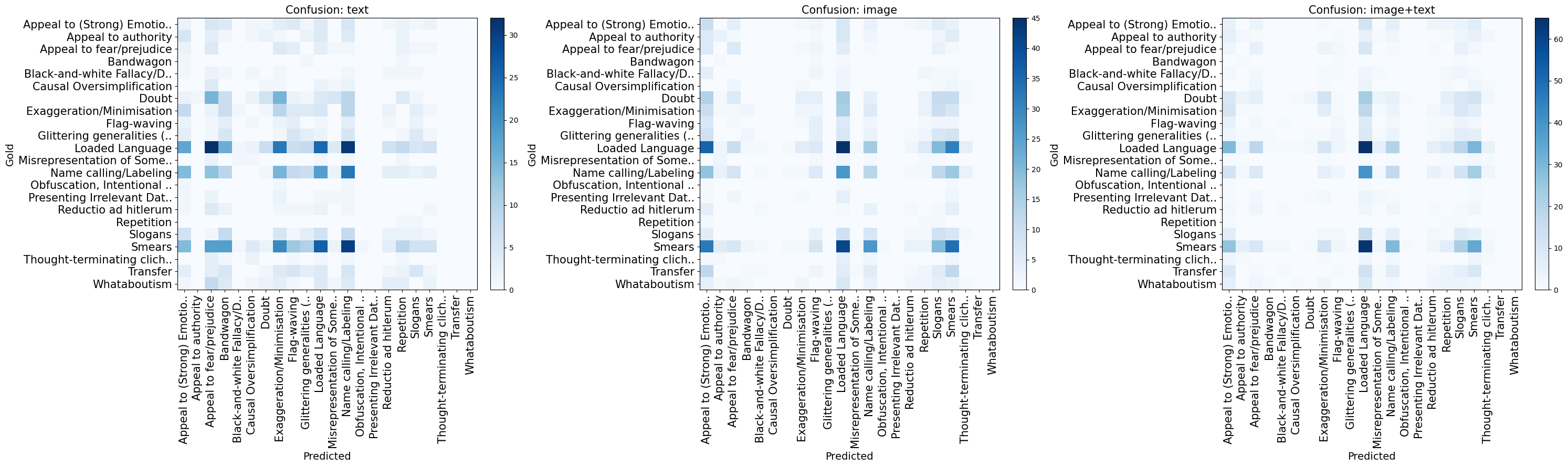}
    \caption{Confusion matrix for PTM-UA without \textit{No Propaganda} (Qwen3-VL-4B, image+text). Former abstentions are redistributed into partially correct technique labels.}
    \label{fig:cm_ptmua_noprop}
\end{figure}

\twocolumn
\section{Model exclusion from Explainability Study}
\label{exclusion}
GPT-4.1-mini is a closed API model, and its internal activations are not accessible, making LRP technically infeasible. Gemma 3 4B is open-weight, but running LRP over a 4B-parameter model with the full dataset was not computationally feasible within the scope of this study. Both models are therefore included as detection baselines only.

\section{Prompt Templates}
\label{app:prompts}

We use two prompt templates depending on input modality. Both prompts receive the full list of 22 persuasion technique labels plus \texttt{NO\_PROPAGANDA} (when included). For the ablation without \textit{No Propaganda} (Section~\ref{sec:no-prop-ablation}), we remove the \texttt{NO\_PROPAGANDA} label and the clause \textit{"or `NO\_PROPAGANDA' if the content doesn't contain persuasion techniques.''} All outputs are requested in Ukrainian.

\subsection{Image+Text and Text-Only Prompt}
\label{app:prompt-base}

\begin{PromptBlock}
You are an expert in analyzing communication 
strategies in visual media.
Classify the provided content into one or 
multiple of the following persuasion techniques:
{unique_labels}
As output, provide JSON with three fields: 
"labels", "referred_words", and 
"referred_image_concepts".
- "labels": list of persuasion techniques used,
  or "NO_PROPAGANDA" if the content doesn't 
  contain persuasion techniques.
- "referred_words": list of the most important 
  INDIVIDUAL words from the content text 
  supporting your classification. Only single 
  words, not phrases. Only key words most 
  relevant to the classification.
- "referred_image_concepts": up to 5 visual 
  elements most strongly supporting your 
  classification, ranked by importance. No 
  duplicates. Empty list if no image. The list 
  could be empty if the image doesn't provide 
  any clues.
Visual elements rules:
- Be concrete: name actual people, objects, 
  symbols. Do NOT write abstract scene 
  descriptions like "people in a stressful 
  situation".
- For visual elements: use the shortest 
  identifying phrase -- "Obama"/"Обама", 
  "flag"/"прапор". Do not use words on the 
  image themselves.
Language note: All values in referred_words and 
referred_image_concepts must be in Ukrainian.
JSON format output example:
{"labels": ["label1", "label2"], 
 "referred_words": ["word1", "word2"], 
 "referred_image_concepts": ["concept1", 
 "concept2"]}
Text from the content: {text}
Output:
\end{PromptBlock}

\subsection{Image-Only Prompt}
\label{app:prompt-image}

\begin{PromptBlock}
You are an expert in analyzing communication 
strategies in visual media.
Classify the provided content IMAGE into one or 
multiple of the following persuasion techniques:
{unique_labels}
As output, provide JSON with two fields: 
"labels" and "referred_image_concepts".
- "labels": list of persuasion techniques used,
  or "NO_PROPAGANDA" if the content doesn't 
  contain persuasion techniques.
- "referred_image_concepts": up to 5 visual 
  elements most strongly supporting your 
  classification, ranked by importance. No 
  duplicates. The list could be empty if the 
  image doesn't provide any clues.
Visual elements rules:
- Be concrete: name actual people, objects, 
  symbols. Do NOT write abstract scene 
  descriptions like "people in a stressful 
  situation".
- For words/text visible in the image: do not 
  include ANY of them to 
  referred_image_concepts, as they belong to 
  the text modality. Only visual elements 
  should be included.
- For visual elements: use the shortest 
  identifying phrase -- "Obama"/"Обама", 
  "flag"/"прапор". Do not use words on the 
  image themselves.
Language note: All values in 
referred_image_concepts must be in Ukrainian.
JSON format output example:
{"labels": ["label1", "label2"], 
 "referred_image_concepts": ["concept1", 
 "concept2"]}
Output:
\end{PromptBlock}

\subsection{Propaganda Technique Labels}
\label{app:labels}

The following 23 labels (including \texttt{NO\_PROPAGANDA}) are provided to the model as \texttt{\{unique\_labels\}}: Appeal to (Strong) Emotions, Appeal to Authority, Appeal to Fear/Prejudice, Bandwagon, Black-and-white Fallacy/Dictatorship, Causal Oversimplification, Doubt, Exaggeration/Minimisation, Flag-waving, Glittering Generalities (Virtue), Loaded Language, Misrepresentation of Someone's Position (Straw Man), Name Calling/Labeling, Obfuscation/Intentional Vagueness/Confusion, Presenting Irrelevant Data (Red Herring), Reductio ad Hitlerum, Repetition, Slogans, Smears, Thought-terminating Clich\'{e}, Transfer, Whataboutism, and NO\_PROPAGANDA.

\section{Category Annotation Prompt}
\label{app:category-prompt}

The following prompt is used with GPT-4o to categorize attributed words. The placeholders \texttt{\{meme\_text\}}, \texttt{\{pred\_labels\}}, and \texttt{\{word\_list\}} are filled per sample.

\begin{PromptBlock}
You are analyzing memes. The meme text is in 
Ukrainian. Categorize words in context.
Meme text: "{meme_text}"
Predicted persuasion techniques: {pred_labels}
The following words were flagged as important 
in this meme by automated methods:
{word_list}
For each word, assign ONE category based on 
its role in THIS specific meme:
- ENTITY: named people, places, organizations, 
  brands
- EMOTIONAL: emotionally loaded, 
  fear/hate/outrage-inducing language
- RHETORIC: calls to action, imperatives, 
  slogans, persuasive framing
- CONTENT: neutral factual nouns, events, 
  dates, statistics
- OTHER: numbers alone, punctuation artifacts, 
  unclear
Return ONLY a JSON object mapping each word 
to its category.
No markdown, no explanation.
Example: {"word1": "ENTITY", 
  "word2": "EMOTIONAL", "word3": "CONTENT"}
\end{PromptBlock}

\section{Hierarchical F1 Definitions}
\label{app:metrics}

Given a taxonomy tree $T$, let $\hat{Y}_i$ and $Y_i$ denote the predicted and gold label sets for sample $i$, and let $\uparrow\!Y$ denote the extension of $Y$ to include all ancestor nodes in $T$ (excluding the root).

\textbf{Per-sample hierarchical precision and recall:}
\begin{equation}
hP_i = \frac{|\uparrow\!\hat{Y}_i \cap \uparrow\!Y_i|}{|\uparrow\!\hat{Y}_i|}, \quad
hR_i = \frac{|\uparrow\!\hat{Y}_i \cap \uparrow\!Y_i|}{|\uparrow\!Y_i|}
\end{equation}

\textbf{Per-sample hierarchical F1:}
\begin{equation}
hF1_i = \frac{2 \cdot hP_i \cdot hR_i}{hP_i + hR_i}
\end{equation}

\textbf{Mean per-sample hierarchical F1:}
\begin{equation}
\overline{hF1} = \frac{1}{N} \sum_{i=1}^{N} hF1_i
\end{equation}

\textbf{Micro-averaged hierarchical F1} (official SemEval metric):
\begin{equation}
\begin{aligned}
hP_{\mu} &= \frac{\sum_{i=1}^{N} |\uparrow\!\hat{Y}_i \cap \uparrow\!Y_i|}
                 {\sum_{i=1}^{N} |\uparrow\!\hat{Y}_i|}, \\
hR_{\mu} &= \frac{\sum_{i=1}^{N} |\uparrow\!\hat{Y}_i \cap \uparrow\!Y_i|}
                 {\sum_{i=1}^{N} |\uparrow\!Y_i|}.
\end{aligned}
\end{equation}

\begin{equation}
hF1_{\mu} = \frac{2 \cdot hP_{\mu} \cdot hR_{\mu}}{hP_{\mu} + hR_{\mu}}
\end{equation}

\textbf{Macro per-label hierarchical F1.} For each node $n$ in the hierarchy $\mathcal{N}$ (all leaves and internal nodes, excluding root), define binary indicators:
\begin{equation}
y_{i,n} = \mathbf{1}[n \in \uparrow\!Y_i], \quad \hat{y}_{i,n} = \mathbf{1}[n \in \uparrow\!\hat{Y}_i]
\end{equation}

Per-node precision, recall, and F1:
\begin{equation}
\begin{aligned}
P_n  &= \frac{\sum_{i=1}^{N} y_{i,n} \hat{y}_{i,n}}
           {\sum_{i=1}^{N} \hat{y}_{i,n}}, \\
R_n  &= \frac{\sum_{i=1}^{N} y_{i,n} \hat{y}_{i,n}}
           {\sum_{i=1}^{N} y_{i,n}}, \\
F1_n &= \frac{2 P_n R_n}{P_n + R_n}.
\end{aligned}
\end{equation}

Macro average over active nodes $\mathcal{N}^* = \{n \in \mathcal{N} : \sum_i y_{i,n} > 0 \lor \sum_i \hat{y}_{i,n} > 0\}$:
\begin{equation}
hF1_{M} = \frac{1}{|\mathcal{N}^*|} \sum_{n \in \mathcal{N}^*} F1_n
\end{equation}

\end{document}